\documentclass{article}

\usepackage{microtype}
\usepackage{graphicx}
\usepackage{subfigure}
\usepackage{booktabs} %
\usepackage{amsfonts}
\usepackage{amsmath}

\usepackage{todonotes}

\usepackage{hyperref}

\newcommand{\rmz}{\mathbf{z}}
\newcommand{\rmx}{\mathbf{x}}

\usepackage[accepted]{icml2021}

\icmltitlerunning{A Study of Joint Graph Inference and Forecasting}

\begin{document}

\twocolumn[
\icmltitle{A Study of Joint Graph Inference and Forecasting}

\icmlsetsymbol{equal}{*}

\begin{icmlauthorlist}
\icmlauthor{Daniel Zügner}{tum,intern}
\icmlauthor{Fran\c{c}ois-Xavier Aubet}{amz}
\icmlauthor{Victor Garcia Satorras}{amst,intern}
\icmlauthor{Tim Januschowski}{amz}
\icmlauthor{Stephan Günnemann}{tum}
\icmlauthor{Jan Gasthaus}{amz}
\end{icmlauthorlist}

\icmlaffiliation{tum}{Technical University of Munich}
\icmlaffiliation{amz}{AWS AI Labs, Amazon Web Services}
\icmlaffiliation{intern}{Work done while being an intern at AWS AI Labs, Amazon Web Services}
\icmlaffiliation{amst}{University of Amsterdam}

\icmlcorrespondingauthor{Daniel Zügner}{zuegnerd@in.tum.de}

\icmlkeywords{Machine Learning, ICML}

\vskip 0.3in 
]

\printAffiliationsAndNotice{}  %

\begin{abstract}
\looseness=-1
    We study a recent class of models which uses graph neural networks (GNNs) to improve forecasting in multivariate time series. 
    The core assumption behind these models is that there is a latent graph between the time series (nodes) that governs the evolution of the multivariate time series. 
    By parameterizing a graph in a differentiable way, the models aim to improve forecasting quality.
    We compare four recent models of this class on the forecasting task. Further, we perform ablations to study their behavior under changing conditions, e.g., when disabling the graph-learning modules and providing the ground-truth relations instead. Based on our findings, we propose novel ways of combining the existing architectures.
\end{abstract} 

\section{Introduction}
Forecasting multivariate time series is a core machine learning task both in science and in industry~\citep{petropoulos2020forecastinglong}. 
Between the individual time series (\emph{nodes}), rich dependencies and interactions (\emph{edges}) govern how the time series evolves. 
In the simplest case these could be (linear) correlations; other examples include the road network underlying traffic flows~\cite{wu2020connecting,shang2021discrete}, or physical relations such as attraction or repulsion affecting trajectories of objects in space~\cite{kipf2018neural}. 

Knowledge of the `true' relations can be used to make more accurate predictions of how the time series evolves in the future, e.g., by using graph neural networks (GNNs) (e.g., \cite{gcn,gat,graphsage,gilmer,scarselli,ppnp,pprgo,netgan}) . 
Even more, the graph can reveal fundamental insights into the system described by the time series, and may thus be of value in itself, independent of an improvement in the forecasting quality. 
Therefore, recent works aim at jointly inferring relations between the time series \emph{and} learn to forecast in an end-to-end manner, sometimes without any prior information about the graph~\cite{wu2020connecting,deng2021graph}.  
Besides potential benefits in forecasting quality, inferring a graph among $N$ time series comes at an inherent computational complexity of $O(N^2)$, which needs to be taken into account when deciding whether to leverage joint graph inference and forecasting. Hence, we consider the following \emph{research questions} in this paper.

\textbf{(R1)} In which scenarios do joint graph inference and forecasting improve forecasting accuracy? Given the diverse domains and settings of multivariate time series forecasting (e.g., underlying spatial relations of sensors in traffic forecasting, sets of sensors measuring different properties of the same system, etc.) it is possible that graph inference helps the forecasting task more in some use cases.

\textbf{(R2)} How do the existing architectures compare in forecasting performance? Are there certain architectural choices that appear beneficial for forecasting?

\textbf{(R3)} What are properties of the inferred graphs by the model? Specifically,  how consistent are the inferred graphs across different training runs? How (dis-)similar are the inferred graphs to the ``ground-truth'' graphs (when known)?

\section{Background}

\textbf{Forecasting with Multivariate Time Series}
\looseness=-1
In time series forecasting we are interested in estimating a future series $\rmz_{t+1:T}$ given its past $\rmz_{t_0:t}$ and some context information about the past $\rmx_{t_0:t}$ where variables $t_0 < t < T$ index over time. For the multivariate case, we can consider $N$ time series at a time $\rmz_{t_0:T} = \{\rmz_{1, t_0:T}, \dots, \rmz_{N, t_0:T}\} \in \mathbb{R}^{N \times T-t_0}$. We model the following conditional distribution:
\begin{equation} \label{eq:prob_model}
p(\rmz_{i,t+1:T} | \rmz_{t_0:t}, \rmx_{t_0,t}), \; 1 \leq i \leq N, 
\end{equation}
where $i$ indexes over time series. Notice that we are conditioning on all $N$ series in order to estimate the series $i$.

\textbf{Time Series Forecasting for graph structured data}

When conditioning over multivariate time series as in Eq.~\eqref{eq:prob_model}, we may benefit from modelling the relations between different multivariate time series. An expressive structure to capture such relations are graphs. We can define a graph as a set of nodes $v_i \in \mathcal{V}$ and edges $e_{ij} \in \mathcal{E}$ that relate the nodes. In our case each $\rmz_i$ is associated to a graph node $v_i$. Edges $e_{ij}$ may be given or unkown depending on the dataset, in cases where the underlying graph is latent/unkown we may jointly infer the graph while estimating a forecasting model. In this work we study the performance of a variety of algorithms under different assumptions of the graph structure (known, unkown, partially known). Note that even in the cases where we have ``ground-truth'' knowledge (e.g., of spatial relations), there may still be additional latent relations which could be discovered by the models.

\section{Literature Review}
\looseness=-1
Recent models perform joint graph learning and forecasting in multivariate timeseries.  These models are \textbf{GTS} (``graph for timeseries'') \cite{shang2021discrete}, Graph Deviation Network (\textbf{GDN}) \cite{deng2021graph}, MTS forecasting with GNNs (\textbf{MTGNN}) \cite{wu2020connecting}, and Neural Relational Inference (\textbf{NRI}) \cite{kipf2018neural}. Here, we briefly introduce these four methods and their differences and commonalities; for a more detailed overview, see Appendix~\ref{app:model_details}. 

\looseness=-1
All models can be decomposed into two main components: the \textbf{graph learning} and the \textbf{forecasting} modules. The former outputs an adjacency matrix describing a graph between the nodes (i.e., timeseries). The latter takes this graph as well as the input timeseries window to forecast the next timestep(s). Once the adjacency matrix has been obtained from the graph learning module, there are many ways of how to leverage it for forecasting the timeseries. The core idea of the models of this study is that the adjacency matrix construction step is differentiable and jointly learned with the forecasting module. Thus, the intuition is that the model will learn graphs which help the forecasting task.

\subsection{Graph learning} The goal of the graph learning module is to output an adjacency matrix $\mathbf{A}\in[0,1]^{N \times N}$, where each entry $\mathbf{A}_{ij}$ denotes the edge weight between nodes $(i,j)$. Typically, we aim for $\mathbf{A}$ to be sparse, which reflects the intuition that there are only relatively few useful relations in the latent graph. Each model first represents each node $i$ by a fixed-size vector $\mathbf{h}_i$, followed by a pairwise similarity computation of any pair $\mathbf{h}_i$ and $\mathbf{h}_j$, e.g., by using a fully connected neural network or simply by taking the dot product. 

\looseness=-1
Next, the models obtain the adjacency matrix from the pairwise scores. \textbf{MTGNN} and \textbf{GDN} do so by taking the $K$ highest scores per node. An advantage of this is that by choosing $K$ appropriately $\mathbf{A}$ is guaranteed to be sparse. On the other hand, the top-$K$ operation is not continuously differentiable, which may pose challenges to end-to-end learning.

\textbf{NRI} and \textbf{GTS} first map the pairwise scores into range $[0,1]$ (e.g., via softmax or sigmoid). The models use the Gumbel softmax trick \cite{maddison2017concrete,jang2017categorical} to sample a discrete adjacency matrix from the edge probabilities in a differentiable way (though gradients are biased); a downside is that we have to take extra steps to obtain a sparse graph, e.g., by regularization.

Moreover, the models can can be broadly split into two groups according to how they compute the fixed-size representations $\mathbf{h}_i$ per node: \textbf{MTGNN} and \textbf{GDN} simply learn these representations as node embeddings; on the other hand, \textbf{NRI} and \textbf{GTS} compute the vectors $\mathbf{h}_i$ based on the time series itself. That is, they apply some (shared) function to each timeseries to map it into a fixed-size vector. While \textbf{NRI} dynamically produces the representations \emph{per individual window}, \textbf{GTS} uses the \emph{whole training timeseries} for each node. The former has the advantage of being more flexible, though more expensive, since we need to compute a $[B \times N \times N]$ tensor to store the \emph{individual} adjacency matrices, where $B$ is the batch size. On the other hand, the graph learned by \textbf{GTS} is global, i.e., shared for all time series. It is thus more efficient yet less flexible, as the model cannot adjust the graph for changing inputs during inference time. Moreover, in its current implementation, this leads to \textbf{GTS}'s number of parameters growing linearly with the length of the training time series (though this could in principle be resolved via dilated convolutions or pooling).

\subsection{Graph-based forecasting}
\looseness=-1
There are many existing models to incorporate graph structure in the forecasting task (e.g., \cite{dcrnn,gcrn,chen2018gc,tgcn,zhu2020a3t,guo2019attention,panagopoulos2020transfer,Wang2018:RNN}). Each of the models in this study has its own way of forecasting the time series given the input timeseries window and the adjacency matrix constructed by the graph learning module.  For instance, \textbf{MTGNN} interchanges temporal convolution layers with graph convolution layers, and \textbf{GTS} uses a Diffusion-Convolutional Recurrent Neural Network (DCRNN) \cite{dcrnn}, where the hidden states of each node are diffused via graph convolutions at each timestep. Again, the core idea is that the adjacency matrix used in the graph-based forecasting is itself constructed in a differentiable way and can thus be adjusted by the model to improve forecasting results.

\section{Experiments}

\begin{table*}
    \centering
     \resizebox{0.7 \textwidth}{!}{
        \begin{tabular}{l|rr|rr|rr|rr}
            \toprule
            {} & \multicolumn{2}{c|}{GTS} & \multicolumn{2}{c|}{MTGNN} & \multicolumn{2}{c|}{NRI} & \multicolumn{2}{c}{GDN} \\
            {} &    MAE@12 &   \multicolumn{1}{c|}{$\Delta$}&   MAE@12 &   \multicolumn{1}{c|}{$\Delta$} &    MAE@12 &    \multicolumn{1}{c|}{$\Delta$} &   MAE@12 &   \multicolumn{1}{c}{$\Delta$} \\
            \midrule
            METR-LA   &    $3.74$ &  $+3.27\%$ &   $3.89$ &  $-9.85\%$ &   $7.8$ &   $-7.69\%$&    $4.1$ &  $+4.87\%$ \\
            PEMS-BAY  &    $1.91$ &  $+5.56\%$ &   $1.97$ &  $-6.38\%$ &   $2.13$ &         - &   $2.12$ &  $-2.31\%$ \\
            Diffusion &  $0.0684$ &  $+3.92\%$ &  $0.117$ &  $-5.24\%$ &  $0.0614$ &  $-44.97\%$ &  $0.122$ &  $-7.88\%$ \\
            DAG       &   $0.695$ &  $-0.48\%$ &  $0.697$ &  $-0.40\%$ &   $0.702$ &   $-0.25\%$ &  $0.692$ &  $-3.25\%$ \\
            \bottomrule
            \end{tabular}
            
     }
     \caption{Average forecasting MAE (over five runs) when disabling the graph-learning and forcing the model to use the ground-truth graph. We also show the \emph{percentage change} of the MAE ($\Delta$); e.g., $-4\%$ means error is reduced by $4\%$ over the base scenario.}
     \label{tab:change_ground_truth}

\end{table*}
\begin{table*}
	\centering
	\resizebox{1.0\textwidth}{!}{
		\begin{tabular}{l|rr|rr|rr|rr||rr|rr}
			\toprule
			{} &  \multicolumn{8}{c||}{Random graph} & \multicolumn{4}{c}{No graph}\\
			\midrule
			{} & \multicolumn{2}{c|}{GTS} & \multicolumn{2}{c|}{MTGNN} & \multicolumn{2}{c|}{NRI} & \multicolumn{2}{c||}{GDN} & \multicolumn{2}{c|}{GTS} & \multicolumn{2}{c}{MTGNN} \\
			{} &    MAE@12 &    \multicolumn{1}{c|}{$\Delta$}  &    MAE@12 &   \multicolumn{1}{c|}{$\Delta$} &    MAE@12 &  \multicolumn{1}{c|}{$\Delta$}   &    MAE@12 &    \multicolumn{1}{c||}{$\Delta$} &    MAE@12 &    \multicolumn{1}{c|}{$\Delta$} &    MAE@12 &    \multicolumn{1}{c}{$\Delta$}\\
			\midrule
			METR-LA       &    $3.77$ &   $+4.10\%$ &    $4.31$ &  $-0.03\%$ &    $7.95$ &   $-5.90\%$ &    $4.05$ &   $+3.68\%$    &    $4.43$ &  $+22.20\%$ &    $4.35$ &   $+0.61\%$\\
			PEMS-BAY      &    $1.86$ &   $+2.86\%$ &     $2.1$ &  $+0.02\%$ &       2.12 &         - &    $2.14$ &   $-1.29\%$   &    $2.15$ &  $+18.93\%$ &    $2.11$ &   $+0.51\%$\\
			WADI          &    $5.97$ &   $+0.80\%$ &    $6.27$ &  $-0.09\%$ &    $7.87$ &   $+3.94\%$ &     $7.2$ &   $-4.32\%$   &    $6.09$ &   $+2.89\%$ &    $6.23$ &   $-1.37\%$\\
			SWaT          &   $0.372$ &  $+24.20\%$ &   $0.684$ &  $-4.81\%$ &   $0.398$ &  $-37.94\%$ &   $0.897$ &  $-17.06\%$   &   $0.574$ &  $+91.76\%$ &   $0.722$ &   $+1.26\%$\\
			Electricity   &   $201.0$ &   $+0.71\%$ &   $185.0$ &  $-1.08\%$ &       - &         - &   $270.0$ &   $-3.70\%$   &   $205.0$ &   $+2.87\%$ &   $198.0$ &   $+5.79\%$ \\
			Solar Energy  &    $2.74$ &   $+3.03\%$ &    $2.71$ &  $+0.83\%$ &       - &         - &     $2.9$ &   $+1.53\%$  &    $3.01$ &  $+13.10\%$ &    $2.78$ &   $+2.96\%$\\
			Traffic       &       - &         - &       - &        - &       - &         - &   $0.014$ &   $+1.92\%$   &       - &         - &  $0.0103$ &  $-23.51\%$\\
			Exchange Rate &  $0.0108$ &   $+9.59\%$ &  $0.0149$ &  $+2.99\%$ &  $0.0121$ &  $+16.37\%$ &  $0.0956$ &  $+11.53\%$ &  $0.0101$ &   $+2.14\%$ &  $0.0145$ &  $-15.75\%$ \\
			DAG           &   $0.697$ &   $-0.22\%$ &     $0.7$ &  $-0.08\%$ &       - &         - &   $0.697$ &   $-2.59\%$  &   $0.698$ &   $-0.02\%$ &     $0.7$ &   $-0.11\%$ \\
			Diffusion     &  $0.0703$ &   $+6.83\%$ &   $0.124$ &  $+0.43\%$ &       - &         - &   $0.132$ &   $-0.16\%$    &   $0.126$ &  $+91.70\%$ &   $0.128$ &   $+4.09\%$\\
			\bottomrule
		\end{tabular}
	}
	\caption{Average forecasting MAE (averaged over five runs) when forcing the model to use a (sparse) random graph (left) or when not using a graph at all in the forecasting (right). Relative performance ($\Delta$) as explained in Fig.~\ref{tab:change_ground_truth}. `-' indicates OOM/timeout after 24 hours.}
	\label{tab:change_random_no_graph}
\end{table*}

To address our research questions \textbf{(R1)}-\textbf{(R3)}, we perform experiments on real-world and synthetic datasets. We repeat all runs five times and report the average; error bars are provided in Table~\ref{tab:forecasting_results} (Appendix).

\subsection{Datasets}
We briefly describe here the datasets that we use; more details can be found in appendix section \ref{app:datasets}. We scale each timeseries to have zero mean and unit variance or to have range $[0,1]$ (only SWaT and WADI, as in \cite{deng2021graph}). For training and evaluation we compute MAE on the \emph{original} scale.

PEMS-BAY and METR-LA \citep{dcrnn} are widely used traffic datasets where we do have knowledge about the underlying graph. To construct the sensor graph, we computed the pairwise road network distances between sensors and build the adjacency matrix using a thresholded Gaussian kernel.

\looseness=-1
We use a range of other multi-variate datasets for which no graph structure is known: Electricity,\footnotemark$^,$\footnotemark \ Solar-energy,\footnotemark$^,$\footnotemark[2] Exchange-rate\footnotemark[2] and Traffic.\footnotemark$^,$\footnotemark[2]
Further, SWaT\citep{swat} and WADI \citep{wadi} are datasets of sensors measuring water-treatment plants. 
In the test split there are annotated anomalies where the creators tampered with the water treatment systems.
Therefore, SWaT and WADI were originally proposed as anomaly detection datasets (and e.g., used in the GDN paper); 
however, since the respective training sets are free of anomalies, we use them for our forecasting experiments.
\footnotetext[1]{\scriptsize{\href{archive.ics.uci.edu/ml/datasets/ElectricityLoadDiagrams20112014}{archive.ics.uci.edu/ml/datasets/ElectricityLoadDiagrams20112014}}}
\footnotetext[2]{\scriptsize{\href{github.com/laiguokun/multivariate-time-series-data}{github.com/laiguokun/multivariate-time-series-data}}}
\footnotetext[3]{\scriptsize{\href{www.nrel.gov/grid/solar-power-data.html}{www.nrel.gov/grid/solar-power-data.html}}}
\footnotetext[4]{\scriptsize{\href{https://pems.dot.ca.gov}{https://pems.dot.ca.gov}}}

\textbf{Synthetic datasets.}
To enhance the real world datasets, we create two synthetic datasets starting with a graph and making sure that the graph has an impact on the connection between the time series. This allows us to speculate that the graph will be of importance for the forecasting of the time series.
We create the \textbf{Diffusion} dataset by using Personalized PageRank (PPR) \cite{ppr} to diffuse the multivariate timeseries.
We create the \textbf{DAG} dataset using a  directed acyclic graph (DAG) and making all the children dimensions be a weighted combination of its parents dimensions.

\subsection{Results}
\textbf{(R1).} Here we analyze the forecasting results on the different datasets at horizons 3, 6, and 12, respectively. For reference, we also add a vanilla LSTM \cite{lstm} baseline that jointly forecasts all timeseries, as well LSTM-U, which consists of $N$ univariate LSTMs. Essentially, the LSTM uses information from \emph{all} timeseries, though lacks typical GNN properties such as permutation equivariance and does not leverage sparsity. LSTM-U is on the other end of the spectrum and simply views all timeseries as completely independent. In Table~\ref{tab:forecasting_results} (appendix) we present the results. 

\looseness=-1
On the popular traffic datasets METR-LA and PEMS-BAY, the GNN models generally dominate the LSTMs. These datasets have known spatial spatial relations among the timeseries, thus this comes as no surprise. NRI's results on METR-LA is quite poor, which we attribute to the relatively large number of nodes and to the fact that the underlying relations are static, while NRI predicts a graph \emph{per window}.

On WADI, interestingly, LSTM-U performs on par with MTGNN. The remaining gap to GTS is relatively small and can potentially be explained by GTS's more sophisticated forecasting procedure. This indicates that on WADI, where we do not have a ``straightforward'' spatial graph between the nodes, the GNN-based models struggle to find useful relations in the data -- or that there are no useful relations in the data to begin with. Similarly, on SWaT, LSTM outperforms all GNN-based models except GTS.

In the synthetic diffusion-based dataset, GTS achieves roughly 50\% lower mean absolute error than LSTM. We attribute this to the fact that the data-generating process (\emph{graph diffusion}) matches well with the DC-RNN architecture used by GTS in the forecasting module. Further, note that on Traffic, GTS ran OOM on a 16GB VRAM GPU for batch size larger than 1, and therefore did not finish within 24 hours. NRI, which is even more computationally expensive, has additional missing values.

In summary, the GNN-based models' edge over the non-graph baselines tends to be largest for datasets with an underlying spatial graph (traffic datasets, Electricity, Solar Energy), and smaller for the datasets where the relations are expected to be more subtle (WADI, SWaT). Future work could compare the GNN-based models to state-of-the-art non-graph forecasting methods in a benchmark study. 

\textbf{(R2).} Next we perform ablation experiments on the GNN-based models to study their behavior when removing their graph-learning module. For the forecasting modules, we either provide the ground-truth graph (where known); provide a sparse random graph; or provide no graph. We compare results to the ``vanilla'' settings of the models, computing the relative change in MAE at horizon 12 in percent.

In Table~\ref{tab:change_ground_truth} we show the results for providing the ground-truth graph to the forecasting modules. Strikingly, MTGNN's performance substantially increases, leading to almost 10\% less MAE on METR-LA. On PEMS-BAY and METR-LA, MTGNN's results are on par with GTS's. This suggests that MTGNN's forecasting module performs well, and that GTS's graph-learning module may be advantageous. GDN also benefits from ground truth, though the effect is not as pronounced. Interestingly, providing the ``true'' graph to GTS leads to a slight performance drop on all but one datasets, indicating that the model's graph-learning module is effective at improving forecasting.

In Table~\ref{tab:change_random_no_graph}, we see the results for providing a (sparse) random Erd\H{o}s Renyi graph to the models (left), or completely disabling the graph processing in the forecasting modules (right). For the random graphs we set the edge probability $p$ such that the expected degree is 30 ($N\geq 100$), 10 ($20\leq N < 100$), or 3 ($N < 20$). An interesting insight is that for GTS, using a random graph has little or moderate effect on most datasets; and that using no graph at all leads to strong performance drop, indicating that GTS's forecasting module greatly benefits from the sparsity of graphs. 

Remarkably, for MTGNN we see relatively little effect when using a random graph or even no graph at all. We hypothesize that this is due to MTGNN's way of constructing the adjacency matrix. It uses kNN-style approach, which has sparse gradients. Further, the edge weights are the result of applying $\mathrm{tanh}$ to the pairwise scores, which may lead to vanishing gradients. Thus, the node embeddings may receive only very little training signal. In contrast, GDN, which also uses node embeddings in the graph learning module, utilizes the node embeddings also in the forecasting task. This may be a way to address the issue of MTGNN. 
Another approach may be to replace the kNN graph construction with differentiable sampling via the Gumbel softmax trick (as in GTS and NRI). This is an interesting experiment to further investigate whether the strategy of parameterizing the graph based on the time series, employed by NRI and GTS, is generally advantageous over node-embedding-based approaches.

\begin{table}
    \centering
\resizebox{0.75 \columnwidth}{!}{
    \begin{tabular}{llrr}
        \toprule
                 &     &  Avg. corr. &  Avg. corr. GT \\
        \midrule
        METR-LA & GDN       &       0.356 &          0.212 \\ %
                 & MTGNN        &      -0.001 &         -0.015 \\
                 & GTS          &       0.264 &         -0.046 \\
                 & GTS w/ reg.  &       0.493 &          0.523 \\
        PEMS-BAY & GDN          &       0.287 &          0.185 \\
                 & MTGNN        &       0.000 &         -0.008 \\
                 & GTS          &       0.164 &         -0.010 \\
                 & GTS w/ reg.  &       0.704 &          0.684 \\
        \bottomrule
        \end{tabular}
    }
\caption{Average correlation of edge scores across different training runs (left), and with the ground-truth graph (right).}
\label{tab:correlations_gt}
\end{table}

\looseness=-1
\textbf{(R3).} Finally, we measure how consistent the learned edge scores are across training runs as well as how similar the learned adjacency matrices are to the ground truth adjacency matrices. For this we measure the \emph{correlation} of edge scores among re-runs and with the ground-truth graph. Intuitively, high correlation means that the model assigns large/small scores to the same node pairs. A subset of the results is shown in Table~\ref{tab:correlations_gt}; see Table~\ref{tab:all_correlation_results} (app.) for more details. We can see that (i) for GDN and GTS, the learned adjacency matrices tend to be moderately similar across training runs. Interestingly, only GDN's learned graphs have a nontrivial correlation with the ground truth. This indicates that the models learn \emph{a} graph which is useful for forecasting, which need not have much in common with the ``true'' (e.g., spatial) graph. Note that for these experiments we have disabled GTS's regularization on the ground-truth graph. When enabling the loss (GTS w/ reg.) we find that, as expected, the learned graphs strongly correlate with the input graph. %

\section{Conclusion}

\looseness=-1
We present a study of recent models performing joint graph inference and forecasting. 
We highlight key commonalities and differences among the architectures. 
In our experiments, we compare the forecasting results of the models and study properties of the different graph-learning modules. For instance, we find MTGNN to be insensitive as to whether the graph-learning module is active or not; though it greatly benefits from access to a ground-truth graph. In general, learning a latent graph is a challenging problem; improvements in terms of expressiveness and computational efficiency could lead to broader applicability.
We highlight potential ways of combining the existing architectures.
\clearpage
\bibliography{references}
\bibliographystyle{icml2021}

\appendix
\clearpage

\begin{table*}
    \centering
    \resizebox{0.75 \textwidth}{!}{

        \begin{tabular}{ll|lll}
            \toprule
            Dataset & Model &  MAE @ 3 &                         MAE @ 6 &                        MAE @ 12  \\
            \midrule
            METR-LA & LSTM &             $3.4950 \pm 0.0104$ &             $3.7121 \pm 0.0117$ &             $4.1049 \pm 0.0113$ \\
                      & LSTM-U &                $3.4164 \pm -$ &                $4.0916 \pm -$ &                $5.1411 \pm - $ \\
                      & NRI &             $4.6802 \pm 0.0813$ &             $6.3878 \pm 0.0873$ &             $8.4661 \pm 0.0985$ \\
                      & GDN &             $3.1490 \pm 0.0165$ &             $3.4818 \pm 0.0144$ &             $3.9096 \pm 0.0123$ \\
                      & MTGNN &             $3.0160 \pm 0.0044$ &             $3.5738 \pm 0.0047$ &             $4.3076 \pm 0.0057$ \\
                      & GTS &    $\mathbf{2.8840} \pm 0.0052$ &    $\mathbf{3.2695} \pm 0.0062$ &    $\mathbf{3.7013} \pm 0.0055$ \\ \hline
            PEMS-BAY & LSTM &             $2.0428 \pm 0.0045$ &             $2.1110 \pm 0.0045$ &             $2.2425 \pm 0.0071$ \\
                      & GDN &             $1.8900 \pm 0.0095$ &             $2.0214 \pm 0.0097$ &             $2.1722 \pm 0.0119$ \\
                      & MTGNN &             $1.3189 \pm 0.0017$ &             $1.6904 \pm 0.0020$ &             $2.1007 \pm 0.0039$ \\
                      & GTS &    $\mathbf{1.2677} \pm 0.0002$ &    $\mathbf{1.5551} \pm 0.0012$ &    $\mathbf{1.8133} \pm 0.0034$ \\ \hline
            WADI & LSTM &             $6.6584 \pm 0.0558$ &             $6.7547 \pm 0.0651$ &             $6.7887 \pm 0.0406$ \\
                      & LSTM-U &             $5.9805 \pm 0.0362$ &             $6.1059 \pm 0.0230$ &             $6.3512 \pm 0.0524$ \\
                      & NRI &                $6.8025 \pm -$ &                $7.0759 \pm -$ &                $7.5703 \pm -$ \\
                      & GDN &             $7.4096 \pm 0.3068$ &             $7.4425 \pm 0.2210$ &             $7.5232 \pm 0.2815$ \\
                      & MTGNN &             $5.9537 \pm 0.0381$ &             $6.1083 \pm 0.0417$ &             $6.2535 \pm 0.0293$ \\
                      & GTS &    $\mathbf{5.4742} \pm 0.0072$ &    $\mathbf{5.5754} \pm 0.0094$ &    $\mathbf{5.7715} \pm 0.0080$ \\ \hline
            SWaT & LSTM &             $0.3000 \pm 0.0133$ &             $0.3296 \pm 0.0105$ &             $0.4324 \pm 0.0163$ \\
                      & LSTM-U &             $0.2869 \pm 0.0012$ &             $0.4868 \pm 0.0008$ &             $0.8826 \pm 0.0011$ \\
                      & NRI &             $0.4147 \pm 0.0139$ &             $0.4798 \pm 0.0142$ &             $0.6408 \pm 0.0110$ \\
                      & GDN &             $0.8029 \pm 0.0447$ &             $0.8537 \pm 0.0689$ &             $1.0812 \pm 0.1495$ \\
                      & MTGNN &             $0.4878 \pm 0.0105$ &             $0.5371 \pm 0.0161$ &             $0.7040 \pm 0.0247$ \\
                      & GTS &    $\mathbf{0.2420} \pm 0.0382$ &    $\mathbf{0.2792} \pm 0.0391$ &    $\mathbf{0.3874} \pm 0.0435$ \\ \hline
            Electricity & LSTM &           $323.3455 \pm 3.8540$ &          $384.2395 \pm 10.7886$ &           $352.4884 \pm 4.2168$ \\
                      & LSTM-U &           $710.9171 \pm 0.7324$ &          $1079.3943 \pm 2.9989$ &           $849.2502 \pm 1.8949$ \\
                      & GDN &           $265.1665 \pm 3.0810$ &           $269.2244 \pm 2.3668$ &           $280.4004 \pm 1.4468$ \\
                      & MTGNN &  $\mathbf{170.1549} \pm 2.9775$ &           $186.0044 \pm 4.7505$ &  $\mathbf{193.4988} \pm 4.4654$ \\
                      & GTS &           $175.8778 \pm 1.0221$ &  $\mathbf{185.7905} \pm 0.9314$ &           $199.5826 \pm 1.3747$ \\ \hline
            Solar Energy & LSTM &             $1.9820 \pm 0.0157$ &             $2.6776 \pm 0.0213$ &             $4.2408 \pm 0.0239$ \\
                      & LSTM-U &             $2.7603 \pm 0.0071$ &             $4.3665 \pm 0.0027$ &             $6.2828 \pm 0.0024$ \\
                      & GDN &             $2.0953 \pm 0.0181$ &             $2.3299 \pm 0.0212$ &             $2.8556 \pm 0.0403$ \\
                      & MTGNN &             $1.5117 \pm 0.0049$ &             $2.0513 \pm 0.0076$ &             $2.6889 \pm 0.0133$ \\
                      & GTS &    $\mathbf{1.4199} \pm 0.0040$ &    $\mathbf{1.9260} \pm 0.0128$ &    $\mathbf{2.6577} \pm 0.0329$ \\ \hline
            Traffic & LSTM &             $0.0157 \pm 0.0002$ &             $0.0178 \pm 0.0003$ &             $0.0173 \pm 0.0003$ \\
                      & LSTM-U &             $0.0285 \pm 0.0001$ &             $0.0332 \pm 0.0002$ &             $0.0289 \pm 0.0001$ \\
                      & GDN &             $0.0132 \pm 0.0001$ &             $0.0134 \pm 0.0001$ &             $0.0137 \pm 0.0001$ \\
                      & MTGNN &    $\mathbf{0.0102} \pm 0.0003$ &    $\mathbf{0.0107} \pm 0.0004$ &    $\mathbf{0.0108} \pm 0.0003$ \\ \hline
            Exchange Rate & LSTM &             $0.0141 \pm 0.0013$ &             $0.0187 \pm 0.0020$ &             $0.0190 \pm 0.0018$ \\
                      & LSTM-U &             $0.0057 \pm 0.0002$ &             $0.0076 \pm 0.0001$ &             $0.0102 \pm 0.0001$ \\
                      & NRI &    $\mathbf{0.0047} \pm 0.0001$ &             $0.0073 \pm 0.0002$ &             $0.0111 \pm 0.0005$ \\
                      & MTGNN &             $0.0109 \pm 0.0023$ &             $0.0146 \pm 0.0033$ &             $0.0136 \pm 0.0013$ \\
                      & GTS &             $0.0047 \pm 0.0000$ &    $\mathbf{0.0070} \pm 0.0000$ &    $\mathbf{0.0099} \pm 0.0000$ \\ \hline
            DAG & LSTM &             $0.6976 \pm 0.0028$ &             $0.7079 \pm 0.0030$ &             $0.7507 \pm 0.0026$ \\
                      & LSTM-U &             $0.7154 \pm 0.0007$ &             $0.7278 \pm 0.0007$ &             $0.7816 \pm 0.0007$ \\
                      & NRI &             $0.6132 \pm 0.0005$ &             $0.6297 \pm 0.0009$ &             $0.7034 \pm 0.0013$ \\
                      & GDN &             $0.6363 \pm 0.0015$ &             $0.6519 \pm 0.0014$ &             $0.7154 \pm 0.0011$ \\
                      & MTGNN &             $0.6107 \pm 0.0005$ &             $0.6277 \pm 0.0009$ &             $0.6999 \pm 0.0010$ \\
                      & GTS &    $\mathbf{0.6088} \pm 0.0003$ &    $\mathbf{0.6254} \pm 0.0004$ &    $\mathbf{0.6960} \pm 0.0004$ \\ \hline
            Diffusion & LSTM &             $0.1064 \pm 0.0001$ &             $0.1355 \pm 0.0002$ &             $0.1405 \pm 0.0003$ \\
                      & LSTM-U &             $0.0986 \pm 0.0000$ &             $0.1333 \pm 0.0000$ &             $0.1393 \pm 0.0001$ \\
                      & NRI &                $0.0704 \pm -$ &                $0.0894 \pm -$ &                $0.1120 \pm -$ \\
                      & GDN &             $0.0890 \pm 0.0005$ &             $0.1109 \pm 0.0003$ &             $0.1325 \pm 0.0003$ \\
                      & MTGNN &             $0.0739 \pm 0.0002$ &             $0.0969 \pm 0.0003$ &             $0.1239 \pm 0.0008$ \\
                      & GTS &    $\mathbf{0.0620} \pm 0.0001$ &    $\mathbf{0.0655} \pm 0.0002$ &    $\mathbf{0.0706} \pm 0.0003$ \\
            \bottomrule
            \end{tabular}
            
    }
    \caption{Results overview. $\pm$ indicates standard error of the mean (SEM) over five runs. Missing rows indicate OOM/timeout after 24 hours.  `$-$' for SEM means that only one run finished witin 24 hours.}
    \label{tab:forecasting_results}
\end{table*}
\clearpage
\section{Additional results}
In Table~\ref{tab:forecasting_results}, we provide the results on the forecasting task. In Table~\ref{tab:all_correlation_results}, we provide additional correlation results of the learned graphs.

\begin{table}
    \begin{tabular}{llrr}
        \toprule
        Model & Dataset &  Avg. corr. &  Avg. corr. GT \\
        \midrule
        NRI & METR-LA &        0.44 &          -0.24 \\
            & WADI &       -0.11 &            n/a \\
            & SWaT &        0.10 &            n/a \\
            & Exchange Rate &        0.68 &            n/a \\
            & DAG &        0.64 &          -0.00 \\
            & Diffusion &        0.85 &           0.02 \\ \hline
        GDN & METR-LA &        0.36 &           0.21 \\
            & PEMS-BAY &        0.29 &           0.19 \\
            & WADI &        0.13 &            n/a \\
            & SWaT &        0.25 &            n/a \\
            & Electricity &        0.17 &            n/a \\
            & Solar Energy &        0.15 &            n/a \\
            & Traffic &        0.10 &            n/a \\
            & Exchange Rate &        0.44 &            n/a \\
            & DAG &        0.22 &           0.04 \\
            & Diffusion &        0.68 &           0.00 \\ \hline
        MTGNN & METR-LA &       -0.00 &          -0.01 \\
            & PEMS-BAY &       -0.00 &          -0.01 \\
            & WADI &        0.00 &            n/a \\
            & SWaT &        0.01 &            n/a \\
            & Electricity &        0.00 &            n/a \\
            & Solar Energy &        0.00 &            n/a \\
            & Traffic &       -0.00 &            n/a \\
            & Exchange Rate &        0.15 &            n/a \\
            & DAG &        0.00 &           0.00 \\
            & Diffusion &       -0.00 &           0.00 \\ \hline
        GTS & METR-LA &        0.26 &          -0.05 \\
            & PEMS-BAY &        0.16 &          -0.01 \\
            & WADI &        0.46 &            n/a \\
            & SWaT &        0.51 &            n/a \\
            & Solar Energy &        0.02 &            n/a \\
            & Exchange Rate &        0.38 &            n/a \\
            & DAG &        0.46 &           0.02 \\
            & Diffusion &        0.01 &          -0.00 \\
        \bottomrule
        \end{tabular}

    \caption{Average correlation of edge scores among different training runs (left), and average correlation of the resulting edge scores with the ground-truth graph (where available).}
    \label{tab:all_correlation_results}
\end{table}

\section{Model details}
\label{app:model_details}
\subsection{Graph for Time Series (GTS)}
GTS (``graph for time series'') \cite{shang2021discrete} is a recent model aiming to jointly learn a latent graph in the time series and use it for MTS forecasting. 
The model consists of two main components: graph learning, and graph-based forecasting. 

\textbf{Graph learning.} The graph learning module first maps the training partition of each time series $\mathbf{z}_i$ to a fixed-size vector representation $\mathbf{h}_i$ via a 1D-convolutional neural network.
Then, all pairs $(\mathbf{h}_i, \mathbf{h}_j)$ are processed by an MLP to output the probability of an edge between $i$ and $j$.
\begin{equation*}
    \mathbf{h}_i = f_{\textnormal{FC}}\left(\mathrm{Vec}\left (f_{\textnormal{conv}}(\mathbf{z}_{i}) \right ) \right), \hspace{0.3cm} \theta_{ij} = \sigma(g_{\textnormal{FC}}\left([\mathbf{h}_i || \mathbf{h}_j]\right)),
\end{equation*}
where $\sigma(\cdot)$ denotes the logistic sigmoid function and $||$ denotes vector concatenation. Finally, a discrete adjacency matrix $\mathbf{A}$ is obtained via element-wise, differentiable sampling, i.e., $\mathbf{A}_{ij} \sim \mathrm{Ber}(\theta_{ij})$, using the Gumbel softmax trick \cite{jang2017categorical,maddison2017concrete}. Note that GTS uses the complete training partition to parameterize the adjacency matrix at each batch. This means that (i) the model can use information from the whole (training) time series at training time; (ii) the model cannot adjust the learned graph at test time; (iii) the number of parameters grows linearly with the length of the input timeseries, which could be improved by adding dilation or pooling to the convolutional encoder.

\textbf{Forecasting.}  The forecasting module uses the graph produced by the graph learning module, a Diffusion-Convolutional RNN (DCRNN) \cite{dcrnn}.
A DCRNN essentially updates hidden states collectively for all series via a graph convolution, which replaces the usual multiplication with a weight matrix.
Thus, \emph{at each time step}, the RNN uses the learnt graph to locally average the hidden representation of timeseries in their graph neighborhood:
\begin{align*}
    \mathbf{R}_{t}&=\sigma\left( \textnormal{GNN}_{R}\left( [\mathbf{Z}_{t} || \mathbf{H}_{t-1}]; \mathbf{A} \right) + b_{R} \right)\\
    \mathbf{C}_{t}&=\mathrm{tanh}\left( \textnormal{GNN}_{C}\left( [\mathbf{Z}_{t} || \mathbf{R}_{t}\odot \mathbf{H}_{t-1}]; \mathbf{A} \right) \textnormal{+} b_{C} \right) \\
    \mathbf{U}_{t}&=\sigma\left( \textnormal{GNN}_{U}\left( [\mathbf{Z}_{t} || \mathbf{H}_{t-1}]; \mathbf{A} \right) + b_{U} \right) \\
    \mathbf{H}_{t}&=\mathbf{U}_{t} \odot \mathbf{H}_{t-1} +(1-\mathbf{U}_{t})\odot \mathbf{C}_t\\
    \textnormal{GNN}(\mathbf{Z}; \mathbf{A})&=\sum_{k=1}^{K}\left(\mathbf{D}_{O}^{-1}\mathbf{A} \right)^k \mathbf{Z} \mathbf{W}_{O}^{(k)} \textnormal{+} \left(\mathbf{D}_{I}^{-1}\mathbf{A} \right)^k \mathbf{Z}  \mathbf{W}_{I}^{(k)}
\end{align*}
where $\mathbf{D}_{O}$ and $\mathbf{D}_{I}$ are diagonal matrices whose entries are the out- and in-degrees of the nodes in $\mathbf{A}$, respectively; 
$\mathbf{W}_{O}^{(k)}$ and $\mathbf{W}_{I}^{(k)}$, $1 \leq k \leq K$ are learnable weight matrices. GTS uses $K=2$.

\textbf{Regularization.} The authors propose to incorporate potential a-priori knowledge about the ground-truth graph via a regularization loss in the form of element-wise binary cross entropy loss between the learned and prior graph.

\subsection{MTS Forecasting with GNNs (MTGNN)}
Like GTS, MTGNN \cite{wu2020connecting} also consists of a graph learning and forecasting module, though the implementations of these modules are different. 

\textbf{Graph learning.} In contrast to GTS, which parameterizes the representation $\mathbf{h}_i$ of a timeseries using a neural network based on the input timeseries, MTGNN learns two embedding vectors per node (i.e., timeseries), i.e., two embedding matrices $\mathbf{E}_1$, $\mathbf{E}_2$. 
Pairwise scores $\mathbf{A}_{ij}$ are computed as
\begin{align*}
    \mathbf{M}_1 &= \mathrm{tanh}(\alpha \mathbf{E}_1 \mathbf{W}_1)\\
    \mathbf{M}_2 &= \mathrm{tanh}(\alpha \mathbf{E}_2 \mathbf{W}_2)\\
    \mathbf{A} &= \mathrm{ReLU}\left(\mathrm{tanh}\left( \alpha \left( \mathbf{M}_1 \mathbf{M}_2^T - \mathbf{M}_2 \mathbf{M}_1^T  \right) \right)\right),
\end{align*}
where the formulation of $\mathbf{A}$ ensures that it is asymmetric, i.e., if $\mathbf{A}_{ij}$ is positive, $\mathbf{A}_{ji}$ is zero.
Finally, only the top $K$ scores per row $\mathbf{A}_i$ are kept to ensure sparsity of the adjacency matrix. The node embeddings are trained in an end-to-end fashion on the forecasting task.

\textbf{Forecasting.} The main difference to the RNN-based forecasting module in GTS is that MTGNN uses temporal convolutions combined with graph convolution layers. 
MTGNN stacks three blocks of interchanging inception-style temporal convolution layers and graph convolution layers to predict the next timestep(s) of the timeseries.

\subsection{Graph Deviation Network (GDN)}
Graph Deviation Network (GDN) \cite{deng2021graph} is a recent model aimed at \emph{anomaly detection} in multivariate timeseries. 
The model is trained on MTS forecasting, and anomalies are flagged when the predicted value deviates strongly from the observed value. 
Since the model is essentially a forecasting method by construction, we chose to include it as a baseline in this work.

\textbf{Graph learning.} Similar to MTGNN, GDN infers the graph by learning a node embedding $\mathbf{v}_i$ per node. 
Specifically, the model builds a k-NN graph where the similarity metric is the cosine of a pair of embeddings.

\textbf{Forecasting.} The forecasting module is based on the Graph Attention Network (GAT) \cite{gat} architecture.
\begin{equation*}
    \mathbf{h}_{i}=\mathrm{ReLU}\left(\sum_{j \in \mathcal{N}(i) \cup \{i\}} \alpha_{ij} \mathbf{W}\mathbf{z}_{j} \right),
\end{equation*}
where $\mathbf{z}_j \in \mathbb{R}^{w}$ is the input timeseries window of node $j$, $\mathbf{W}$ is a learned weight matrix, and $\alpha_{ij}$ are attention scores, computed as follows.
\begin{align*}
     \alpha_{ij}=\mathrm{softmax}_j\left( \mathrm{LeakyReLU}\left(\mathbf{a}^T \left[ \mathbf{g}_{i} || \mathbf{g}_j \right]\right) \right),
\end{align*}
where $\mathbf{g}_{i} = [\mathbf{v}_i || \mathbf{W}\mathbf{z}_{i}]$, and $\mathbf{a}$ is a learned vector. 
The next predicted value(s) for all nodes are predicted jointly by a stack of fully connected layers $f_\textnormal{MLP}$:
\begin{equation*}
    \hat{\mathbf{s}} \in \mathbb{R}^{N \times \hat{T}} = f_\textnormal{MLP}\left(\left[ \mathbf{v}_{1} \odot \mathbf{h}_1 || \mathbf{v}_{2} \odot \mathbf{h}_2 || \ldots || \mathbf{v}_{N} \odot \mathbf{h}_N    \right]\right),
\end{equation*}
where $\hat{T}$ is the number of predicted timesteps.

\subsection{Neural Relational Inference (NRI)}
The Neural Relational Inference (NRI) \cite{kipf2018neural} model assumes a slightly different setting than GTS, MTGNN, and GDN. 
Instead of learning a \emph{global, static} graph over the whole time series, NRI infers a graph \emph{per input window}. 
While this setting is more flexible, it comes at the drawback of having inherent $O(B \times N^2)$ memory complexity, where $B$ is the batch size. 
Thus, NRI can only realistically scale to small graphs, i.e. at most $N\approx 50$. NRI is a VAE-based architecture consisting of an encoder module predicting edge probabilities and a decoder module performing the forecasting.

\textbf{Graph learning.} The encoder module interchanges neural networks on the node and edge representations, respectively:
\begin{align*}
    \mathbf{h}_j^{(1)} &= f_{\textnormal{emb}}(\mathbf{z}_j)  &\mathbf{h}_{(i,j)}^{(1)} = f_{\textnormal{e}}^{(1)}([\mathbf{h}_{i}^{(1)} || \mathbf{h}_{j}^{(1)}]) \\
    \mathbf{h}_{j}^{(2)} &= f_{\textnormal{v}}^{(1)}(\sum_{i\neq j} \mathbf{h}^{(1)}_{(i,j)}) &\mathbf{h}_{(i,j)}^{(2)} = f^{(2)}_{\textnormal{e}}([\mathbf{h}^{(2)}_{i} || \mathbf{h}^{(2)}_{j}])
\end{align*}
Finally, the edge type posterior is $q_{\phi}(\mathbf{z_{(i,j)}}|\mathbf{z_{t_0:t}})=\mathrm{softmax}(\mathbf{h}^{(2)}_{(i,j)})$, where one edge type can be hard-coded to denote `no edge'. 
Similar to GTS, NRI samples a discrete graph using the Gumbel softmax trick.

\textbf{Forecasting.} The decoder module is similar to the encoder. It has a fully connected neural network per edge type, which processes the respective input pairs of nodes connected by the specific edge type. 
Finally, for each node, the representations of incoming edges are aggregated, and a final neural network predicts the value of the next timestep. 
For the decoder, the authors propose an MLP-based and a RNN-based variant.

\section{Datasets}
\label{app:datasets}
We describe here more in details the characteristics of the datasets used.

In Table~\ref{tab:real_world} we summarize the real-world datasets used in this work.
SWaT\citep{swat} and WADI \citep{wadi} are datasets of sensors measuring water-treatment plants. 
In the test split there are annotated anoalies where the creators tampered with the water treatment systems.
Therefore, SWaT and WADI were originally proposed as anomaly detection datasets (and e.g., used in the GDN paper); 
however, since the respective training sets are free of anomalies, we use them for our forecasting experiments.

PEMS-BAY and METR-LA \citep{dcrnn} are widely used traffic datasets where we do have knowledge about the underlying graph. To construct the sensor graph, the computed the pairwise road network distances between sensors and build the adjacency matrix using a thresholded Gaussian kernel.

\begin{table}
	\centering
	\begin{tabular}{l|ccc}
		\toprule
		Dataset & \# Samples & $N$ & Ground truth? \\ \midrule
		PEMS-BAY \cite{dcrnn}                         & 52,116 & 325 & Yes \\
		METR-LA \cite{dcrnn}                       & 34,272 & 207 & Yes \\ \midrule
		WADI\footnotemark] \cite{wadi}                             & 1,187,951 & 122 & No  \\
		SWAT\footnotemark[5] \cite{swat}                             & 475,200 &  51 & No  \\
		Electricity\footnotemark$^,$\footnotemark     & 26,304 & 321 & No  \\
		Solar-energy\footnotemark$^,$\footnotemark[7] & 52,560 & 137 & No  \\
		Exchange-rate\footnotemark[7]                 &  7,588 &   8 & No  \\
		Traffic\footnotemark$^,$\footnotemark[7]      & 17,544 & 862 & No  \\ \bottomrule
	\end{tabular}
	\caption{Real-world dataset summary.}
	\label{tab:real_world}
\end{table}
\footnotetext[5]{As proposed by \cite{deng2021graph}, we subsample SWaT and WADI by a factor of 10 in the time dimension using the median operation.}
\footnotetext[6]{\url{archive.ics.uci.edu/ml/datasets/ElectricityLoadDiagrams20112014}}
\footnotetext[7]{\url{github.com/laiguokun/multivariate-time-series-data}}
\footnotetext[8]{\url{www.nrel.gov/grid/solar-power-data.html}}
\footnotetext[9]{\url{https://pems.dot.ca.gov}}

\subsection{Synthetic datasets}
One drawback about real-world datasets -- even the traffic datasets for which we have some knowledge about the relations -- is that we do not know the true data-generating process and how the graph interacts with it. To address this, we generated two synthetic datasets where relations between nodes are handcrafted. An advantage of this is that we know in advance the true dependencies between nodes. %

\textbf{Diffusion-based dataset.} For each of the $N$ timeseries, we first randomly sample parameters of a sinusoidal function, i.e., its frequency, amplitude, horizontal, and vertical shift. Next, we partition the nodes into $K$ clusters and generate an undirected graph from a Stochastic Blockmodel (SBM), such that nodes within a cluster are more densely connected than between clusters. We use Personalized PageRank (PPR) \cite{ppr} to diffuse the multivariate timeseries, i.e., compute for each timeseries the weighted combination of itself and the other nodes in its vicinity. For each node, we add independent Gaussian noise to the other nodes before averaging. Thus far we have induced correlation between nodes in the same cluster. Finally, we perform a weighted combination of the timeseries \emph{before} and \emph{after} diffusion, where the diffused timeseries is lagged by $C$ timesteps. This means, each timestep $\mathbf{z}_t=\alpha \cdot \tilde{\mathbf{z}}_{t} + (1-\alpha)\cdot \hat{\mathbf{z}}_{t-C}$. This way, knowing the value of $i$'s neighbors $C$ steps is useful to predict its current value, rewarding models which have correctly identified the relations.

\textbf{DAG-based dataset.} As an alternative, we generate a dataset based on a directed acyclic graph (DAG). We induce an arbitrary topological order defined by the node IDs, i.e., $1, \ldots, N$. We iterate over nodes in increasing topological order. For node $i$, we randomly sample \emph{incoming} edges from all nodes $j<i$ with uniform probability $p$. If no incoming edges were sampled for $i$, we generate its timeseries as a random sinusoidal function as described above and add Gaussian noise. Otherwise, $i$'s timeseries is a randomly weighted combination of modified timeseries $j$ for which $(i,j)$ is an edge and $j<i$. We modify the input timeseries by applying random horizontal and vertical shift and stretch, and add some noise again. Thus, $i$'s values are directly determined by its incoming edges (except for some noise), and we expect models which correctly identify the relations to perform well in the forecasting task.

For the diffusion dataset we set $\alpha=0.75$, $C=10$, and the restart probability in the PPR computation to $0.15$. We choose $K=5$ clusters; the edge probability within clusters is $0.5$, and between clusters we have $0.05$.

For DAG, the edge probability $p=0.1$.
For both synthetic datasets, we set $N=100$.

\section{Training details}
Generally, we use the hyperparameters provided by the authors of the respective papers. We train all models on a horizon of 12 timesteps, where the training loss is the mean absolute error of all 12 time steps. For all datasets except SWaT and WADI, we ignore targets with value zero (as in \cite{shang2021discrete,wu2020connecting}), as these correspond to missing values.  We train models for a maximum of 200 epochs and use early stopping with patience of 20 epochs. We use the validation MAE for early stopping. For SWaT and WADI as well as DAG and Diffusion, we fix the window length to 20. For METR-LA and PEMS-BAY, we set the window length to 12, as proposed by \cite{wu2020connecting,shang2021discrete}. For Electricity, Solar Energy, Traffic, and Exchange Rate, window size is 168 as in \cite{wu2020connecting}.

\end{document}